\documentclass{llncs}
\usepackage{graphicx}
\usepackage{color}

\usepackage{paralist}

\newcommand{\fn}[1]{#1}
\usepackage{color}
\usepackage{ifthen}
\newboolean{fn}
\setboolean{fn}{false}
\ifthenelse{\boolean{fn}}{\renewcommand{\fn}[1]{\textcolor{blue}{#1}}}{}

\newcommand{\fc}[1]{\textcolor{red}{#1}}
\newboolean{fc}
\setboolean{fc}{true}
\ifthenelse{\boolean{fc}}{\renewcommand{\fc}[1]{{}}}{}
\usepackage{umoline}

\begin{document}

\title{What's in an `is about' link? Chemical diagrams and the Information Artifact Ontology}
\titlerunning{Information ontology and chemistry}

\author{Janna Hastings\inst{1,2}\thanks{To whom correspondence should be addressed, email: hastings@ebi.ac.uk} 
			  \and Colin Batchelor\inst{3}
        \and Fabian Neuhaus\inst{4,5} \and \\Christoph Steinbeck\inst{1}}
\authorrunning{J. Hastings, C. Batchelor, F. Neuhaus, C. Steinbeck}

\institute{Chemoinformatics and Metabolism, \\European Bioinformatics Institute, Cambridge, UK
\and Swiss Center for Affective Sciences, University of Geneva, Switzerland
\and Informatics, Royal Society of Chemistry, Cambridge, UK
\and National Institute of Standards and Technology, Gaithersburg, MD, USA
\and University of Maryland Baltimore County, MD, USA}

\maketitle

\begin{abstract}

The Information Artifact Ontology is an ontology in the domain of information entities.  
Core to the definition of what it is to be an information entity is the claim that an information entity must be `about' something, which is encoded in an axiom expressing that all information entities are about some entity.  
This axiom comes into conflict with ontological realism, since many information entities seem to be about non-existing entities, such as hypothetical molecules.  
We discuss this problem in the context of \fn{diagrams} of molecules, a kind of information entity pervasively used throughout computational chemistry. \fn{We then propose a solution that recognizes that information entities such as diagrams are expressions of diagrammatic languages. In so doing, we not only address the problem of classifying diagrams that seem to be about non-existing entities but also allow a more sophisticated categorisation of information entities.}\fc{\marginpar{ It is not obvious to readers what 'non-represenatational' means, so I rephrased}} 
\end{abstract}

\section*{Introduction}

As the importance of ontology in biomedicine grows, the attention of ontologists is being pressed to the tasks of disambiguation of domain terminology and clarification of underlying hierarchies and relationships in an ever-wider network of interrelated domains \cite{Bodenreider2006,smith2010}. Some issues are emerging as similarly problematic in many of these different domains. One such is the clear definition and distinction of foundational types such as \textit{processes} and \textit{dispositions} \cite{batchelor2010}.  Another is the confusion between \textit{information entities}, such as computer simulations, models and diagrams, and the entities that they are models and diagrams \textit{of}.  It is to this latter problem that we turn in this paper.  

Chemical graphs are the molecular models that are used throughout chemistry to succinctly describe chemical entities and allow for computational manipulations \cite{Trinajstic1992,hastings2010}.  Chemical graphs are typically depicted graphically as schematic illustrations -- chemical diagrams.  Chemical graphs and chemical diagrams are examples of information entities in the chemical domain, and their use has become so pervasive that language used by chemists to refer to chemicals regularly interchanges words for information (such as `graph') with words for actual chemicals \cite{hastings2010}.

The Information Artifact Ontology (IAO) \cite{iaogooglecode} is an ontology being developed for the domain of information entities of relevance in biomedicine.  The fundamental criterion by which information entities are defined and categorised in the IAO is their \textit{aboutness}, that is, the types of entities that they are \textit{about}.  A diagram illustrating the chemical structure of caffeine molecules, for example, is about the class of caffeine molecules.  While in this case the chemical diagram corresponds to something in reality (caffeine molecules), there are many other useful and scientifically relevant chemical diagrams that are not about something that exists. Thus, these chemical graphs are not information entities as currently defined in IAO. A similar scenario applies to many other models used in biomedicine, for example pathway diagrams and the mathematical models used in quantitative systems biology. Using chemical diagrams as examples, we will argue that information entities in IAO are defined too narrowly. Since information entities may not necessarily be about something, they cannot be categorized merely by what they are about. But, as we will argue, they should rather be categorised by what sort of information entities they are in their own right. 

The remainder of this paper proceeds as follows.  In the next section we briefly describe the IAO and the theory of chemical graphs and their related diagrams.  Thereafter, we highlight the insufficiency of aboutness in defining types of diagrams.  We go on to introduce some semantics for the representation relationship between chemical diagrams and chemical entities; and finally, we propose a modified approach to information ontology that is free of the problems with the current approach.

\section{Background}

\subsection{The Information Artifact Ontology}

The Information Artifact Ontology (IAO) \cite{iaogooglecode} is an ontology of information entities being developed in the context of the \fn{Open Biological and Biomedical Ontologies} (OBO) Foundry \cite{Smith2007}, beneath the upper level ontology \fn{Basic Formal Ontology} (BFO) \cite{Smith04thecornucopia,Grenon04biodynamicontology}.  Within this context, information entities are defined as:

\begin{definition}
An information content entity (\emph{ICE}) is an entity that is generically dependent on some artifact and stands in the relation of \emph{aboutness} to some entity.
\end{definition}

The generic dependence on an artifact (i.e., a human creation) in the above definition restricts the scope of the domain to human-created information entities. The `generic' part of the dependence captures the intuition that information can be copied, that is, reproduced in multiple bearers, in a way that hair colour, for example, cannot.  
The textual definition also refers to a relation of `aboutness', which is further supplemented by the \fn{axiom}:
\begin{equation}
 \textit{ICE} \texttt{ subClassOf } \textbf{is about} \texttt{ some } \textit{Entity}
\label{eq:iceabout}
\end{equation}

The above is given in the Manchester Web Ontology Language (OWL) syntax, in which the existential quantification ($\exists$) is expressed using the infix \texttt{some} operator.  This should not, however, obscure the strong existential dependency claimed, namely: for every \textit{ICE}, there exists some entity to which the \textit{ICE} is related by the \textbf{is about} relationship. 

\begin{figure}[h]
\centering
		\includegraphics[scale=0.65]{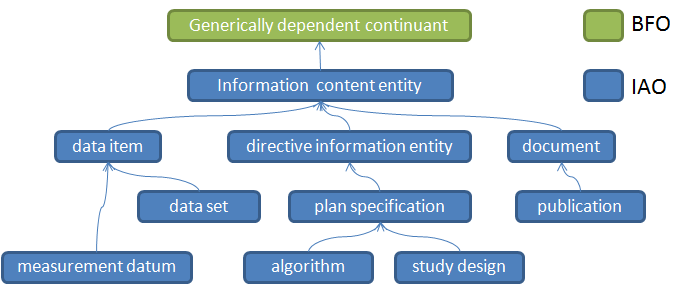}
\caption{An overview of the Information Artifact Ontology}
\label{fig:IAOContent}
\end{figure}

A hierarchical overview of the IAO together with some examples of information content entities (\textit{ICE}s) is illustrated in Figure \ref{fig:IAOContent}.

\subsection{Chemical graphs and diagrams}

The principal object of graph theory is a graph, which consists of a set of objects and the binary relations between them. Graph theory has found many applications in chemistry and is used to represent molecular entities through the molecular graph. These graphs represent the constitution of a molecule in terms of nodes (usually atoms, but in some cases groups of atoms) and edges (chemical bonds) \cite{Trinajstic1992}. 

For the purposes of this paper we define chemical graphs as follows\footnote{We ignore additional complexity such as the representation of stereochemistry.}. 

\begin{definition}
A \emph{chemical graph}, denoted \emph{CG}, is a tuple $(V,E)$ in which each vertex $ i \in V $ corresponds to an atom in a molecule; and each undirected edge $\{ i, j\} \in E $ corresponds to a chemical bond between the atoms $i$ and $j$. 
\end{definition}

These \textit{CG}s are based on the valence bond model of quantum mechanics \cite{Pauling1928}. For many of the molecules most relevant to the pharmaceutical industry this model reasonably accurately represents (1) by atoms, those portions of the molecules that chemists associate with particular atoms, and (2) by bonds, those portions of the molecules that have high electron probability density. Cheminformatics software uses these to make useful predictions about the chemical properties of a molecule so represented and the physical properties of an ensemble of those molecules. They also enable the schematic representation of molecules in diagrams.

\begin{definition}
A \emph{chemical diagram}, denoted \emph{CD}, is a diagrammatic illustration of the information encoded in a \emph{CG}, which follows an agreed \emph{diagrammatic syntax} for the representation of the graph information. 
\end{definition}

\begin{figure}[htp]
\centering
	\includegraphics[scale=0.70]{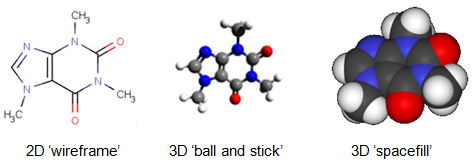}
\caption{Some examples of CDs for the molecule caffeine}
\label{fig:FigChemGraph}
\end{figure}

Some examples of \textit{CD}s are illustrated in Figure \ref{fig:FigChemGraph}.  In the 2D wireframe depiction, the diagrammatic syntax used specifies that the \textit{CD} corresponds to the \textit{CG} in that, for each edge $\{ i, j\} \in E $ there is a corresponding line, and for each vertex $ i \in V $  there is a corresponding \textit{corner} or \textit{line ending} in the \textit{CD}. In the 3D ball and stick diagram, edges are illustrated with lines while vertices are illustrated with coloured, labelled spheres.  In the 3D spacefill diagram, vertices are illustrated with large coloured spheres. Both the colours and the radii of the sphere are arbitrary---atoms are much too small to have colours, but the radii are based on experimental averages and are an approximation to the actual molecular structure.  

Notice that there is not a one-to-one correspondence between \textit{CD}s and \textit{CG}s, since the same \textit{CG} can be illustrated in many different \textit{CD}s, obeying different syntaxes. 

\textit{CD}s, like maps, represent \textit{spatial} information.  Let us call spatial representations such as street maps, chemical diagrams, and engineering design models \textit{structural diagrams} and, to a first approximation, assume that they have a direct structural association with a portion of reality, which they are intended to represent.

\begin{definition} 
A structural diagram (\emph{SD}) is a diagrammatic representation of spatial aspects, such as position, topology and connection, of a structured portion of reality.
\label{def:strawman}
\end{definition}

This definition, however, does not suffice, for reasons that will be described in the following section.

\section{When `is about' isn't enough}

The agreed syntax of \textit{CD}s allows their informational content to be reliably understood by all members of the community who use them for exchange of such information.  


The agreed syntax also allows for the depiction of molecules, which are
\begin{compactenum}
	\item Planned, in that the representation is used as a precursor to a synthesis procedure expected to produce a corresponding molecule instance. 
	\item Hypothesised, in that the representation corresponds to a molecule class for which it is not known whether corresponding instances exist.
	\item Chemically infeasible, in that it is known that the representation illustrates a class of molecules for which no instances can exist for a measurable duration of time under normal conditions.  
	\item Impossible, in that the representation \textit{cannot} be the structure of any molecule instances, since it violates the rules of molecular compositionality.
\end{compactenum}

In the first two cases the \textit{CD} might or might not be \textit{about} molecules that exist.  In the third case chemists expect, and in the fourth case they are certain that, the aboutness criterion of the IAO is violated. Nevertheless, these \textit{CD}s are used by chemists to communicate and exchange information in the same ways as \textit{CD}s that are known to correspond to something in reality.  Thus, the way \textit{CD}s are used does not justify treating only a subset of them as information entities.  It also indicates that Definition \ref{def:strawman} is not along the right lines.  

\begin{figure}[htp]
\centering
	\includegraphics[scale=0.65]{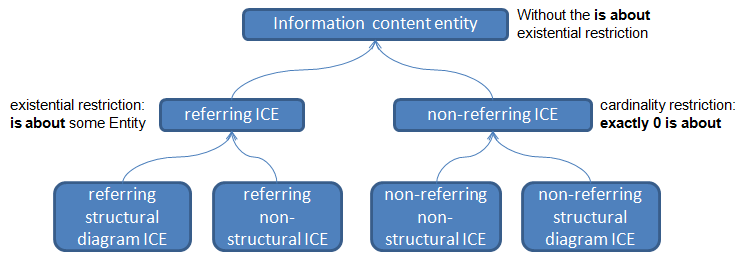}
\caption{Referring and non-referring information entities in the IAO}
\label{fig:IAOPlusNonReferring}
\end{figure}

A conceptualist resolution to this issue might defend a view of ontology as containing representations of \textit{concepts}, and thereby not be required to differentiate between chemical diagrams for real or impossible molecules, or differentiate at the level of metadata only \cite{dumontier2010}. However, this seems to overlook the fundamental distinction between these cases, one that chemists recognise. Another strategy for addressing this problem is provided by Ceusters and Smith \cite{ceusters2010} who distinguish between \textit{referring} and \textit{non-referring} representational units in the context of a mental representation. The application of this distinction to an ontology of \textit{SD}s beneath IAO is illustrated in Figure \ref{fig:IAOPlusNonReferring}.

One obvious problem with this approach is that it leads to a massive level of parallel maintenance, since most types of \textit{ICE} can appear twice in the ontology. A more fundamental objection is that this approach violates the fundamental design principles of BFO: categorization according to \textit{ontological nature}, which does not change. For example, it is impossible for a tree (an independent continuant) to become a temporal region, or for a smile (a dependent continuant)  to become a soccer game (an occurrent).  However, according to the approach in \cite{ceusters2010} a \textit{CD} might be a non-referring \textit{ICE} now, but become a referring \textit{ICE} tomorrow, because somewhere in some lab somebody accidentally synthesized the corresponding molecule.  Thus, in contrast to the other ontological categories in BFO, it would be possible for non-referring \textit{ICE}s to change their ontological nature.  Even worse, the ontological nature of \textit{CD}s would be affected by events that had no causal connection to the \textit{CD} and did not change its structure in any way.  Since the ontological nature of an entity is not affected by Cambridge changes, that is to say changes only in its description, we conclude that `non-referring \textit{ICE}' and `referring \textit{ICE}' are not true ontological categories. 

In summary, we agree with Ceusters and Smith that non-referring \textit{ICE}s are \textit{ICE}s. However, we reject the idea that the distinction between referring and non-referring should be the primary basis for classifying \textit{ICE}s.  There are some \textit{ICE}s that are necessarily about something (e.g., photographs).  But structural diagrams are information entities in virtue of the fact that they are well-formed \textit{expressions} in a \textit{diagrammatic language}.  For each type of \textit{SD}, there is a vocabulary (the symbols and icons that are used in diagrams of that type), a grammar that regulates how the elements of the vocabulary can be combined, and \textit{compositionality} in the sense that the semantics of a complex expression is determined by the semantics of its components and the way these components are arranged.

The elements of the vocabulary of the diagrammatic language do need to correspond to something existing, otherwise the diagrams will not be scientifically relevant. However, not all combinations of the vocabulary that are permissible by the grammar will correspond to something in reality. It would seem strange indeed, on giving an ontological account of natural language, to divide all sentences into those that are about facts and those that are not. ``Submariners love periscopes." is a declarative sentence with a transitive verb regardless of whether it is a fact that submariners love periscopes. The same is true for expressions of diagrammatic languages.

\section{The ontology of structural diagrams}

Different types of \textit{CD} (such as 2D wireframe, 3D ball and stick) obey different diagrammatic syntaxes. What is essential to distinguish different types of diagrams is thus to provide a definition for these syntaxes. 

\begin{definition}
A \emph{diagrammatic language} $L_D = \langle V,G \rangle$ is an ordered pair that consists of \fn{the vocabulary $V$ (a set of icons and symbols)} and a syntax $G$ of composition rules.
\end{definition}

\begin{definition}
   An \emph{interpreted diagrammatic language} is a quadruple $IL_D = \langle V,G,T,\phi \rangle$ such that $\langle V,G \rangle$ is a diagrammatic language, $T$ is a set of types that is partitioned set of independent continuants $IT$ and dependent
continuants $DT$, and $\phi$ is a function that maps the elements from $V$ onto $T$.
\end{definition}

\begin{definition} \label{strucdiag}
  Let $IL_D$ be an intepreted diagrammatic language as above, and let $D$ be a well-formed expression in $L_D$ (i.e., a diagram).
  $D$ is a \emph{structural diagram} that \textbf{is about} an entity $x$ iff there is some injective interpretation function $\iota$ such that:
 \begin{compactitem}
		\item for each element of $V$ and each token $t$ of $V$ that is part of $D$, $\iota(t)$ is an instance of $\phi(V)$
		\item for two tokens $t_1$, $t_2$ that are part of $D$ and $\iota(t_1)$, $\iota(t_2)$ are
instances of elements of $IT$: $t_1$ is connected to $t_2$ iff $\iota(t_1)$ is connected to $\iota(t_2)$
		\item for all tokens $t$, $t_1$, ... $t_n$: if $\iota(t)$ is an instance of some element of $DT$ and $t_1$ ... $t_n$ are all connected to $t$,   then $\iota(t)$ inheres in $\iota(t_1)$ ... $\iota(t_n)$.
		\item there is no part $y$ of $x$ such that $y$ is an instance of some type in $T$ and for all $t$ that are part of $D$ there is no $\iota(t) = y$.
 \end{compactitem}
\end{definition}

Chemical diagrams of hypothetical molecules that do not exist are not \textit{about} anything, but they are still well-formed expressions of an interpreted diagrammatic language. \fn{For example, the vocabulary $V$ of the 3D ball and stick language consists of colored spheres and lines. The syntax $G$ describes how these elements can be combined to diagrams. The set $IT$ consists of types of atoms, the set $DP$ consists of the types of chemical bonds that connect atoms within a molecule. The  function $\phi$ maps the color-coded balls to types of atoms and the links to types of bonds. The second diagram in Figure \ref{fig:FigChemGraph} is a structural diagram of a given instance of a caffeine molecule $x$, since it is possible to map the spheres of the diagram to the atoms that are part of $x$ and the links of the diagram to the chemical bonds of $x$ such that the connections in the diagrams corresponds to the chemical reality in the molecule. Conversely, if the diagram contains a link that does not correspond to a bond in a given molecule $x$ or if it contains a sphere that is mapped to a type of atoms that do not occur as part of $x$, then the diagram does not represent $x$.}\footnote{\fn{The second clause of definition \ref{strucdiag}  is irrelevant in the case of \textit{CD}s, because in \textit{CD}s tokens of symbols for independent continuants (the atoms)  are always connected by tokens of symbols for dependent continuants (the bonds). However, definition \ref{strucdiag} is also intended to be applicable to diagrams where symbols for independent continuants might be connected directly; for example architectural drawings and engineering blueprints.}}

To place \textit{SD}s (and therefore \textit{CD}s) as subtypes of IAO's \textit{ICE}, we need to change the fundamental aboutness criterion from Equation~(\ref{eq:iceabout}) to a \textit{value} rather than \textit{existential} restriction: 

\begin{equation}
 \textit{ICE} \texttt{ subClassOf } \textbf{is about} \texttt{ only } \textit{Entity}
\label{eq:iceaboutfixed}
\end{equation}
This restriction no longer expresses an existential dependence.  Rather, it now has the effect that \textit{if} there is some entity that the \textit{ICE} is about, \textit{then} it must be of the required type to avoid a logical inconsitency. Note that this formula expresses a schema, which will be made more precise for different types of \textit{ICE}. With the inclusion of \textbf{conforms to} axioms to relate the \textit{ICE} to the $L_D$, we are now in a position to provide a better definition for \textit{SD}s and \textit{CD}s to replace Definition~\ref{def:strawman}:

\begin{eqnarray*}
 \textit{SD} \texttt{ subClassOf } \textit{ICE} & \texttt{ and } & \textbf{is about} \texttt{ only } \textit{StructuredEntity} \\
 																				& \texttt{ and } & \textbf{conforms to} \texttt{ some } \textit{DiagrammaticLanguage} \nonumber \\ 	
 \textit{CD} \texttt{ subClassOf } \textit{SD} & \texttt{ and } & \textbf{is about} \texttt{ only } \textit{MolecularEntity} 
\label{eq:smabout}
\end{eqnarray*}

\begin{figure}[htp]
\centering
	\includegraphics[scale=0.50]{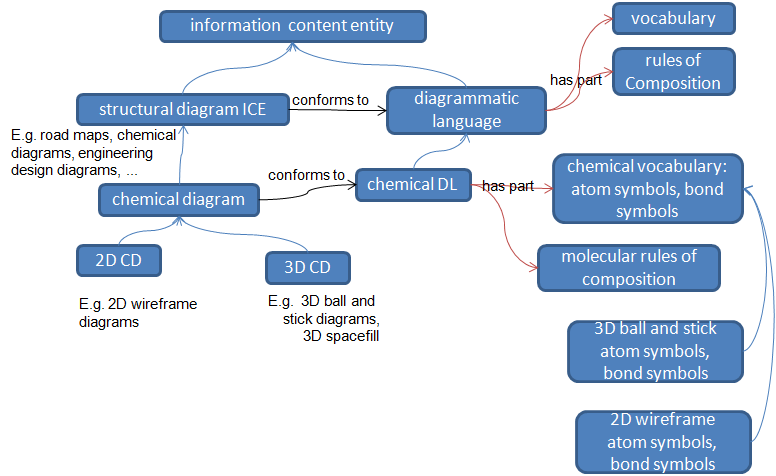}
\caption{The ontology of chemical diagrams with distinctions for different syntaxes}
\label{fig:FigChemDiagOntology}
\end{figure}

We can safely include in the resulting ontology, illustrated in Figure~\ref{fig:FigChemDiagOntology}, diagrams of planned, hypothetical, infeasible, and impossible molecules. 

Now, we can define different types of chemical diagrams regardless of their aboutness, and furthermore express the difference between different \textit{types} of diagrams that are about the same entity (such as 2D and 3D diagrams of caffeine molecules).  However, we can go one step further and define a \textit{relationship} between 2D and 3D depictions of the same molecule. 

\begin{definition}
Let $L_1$, $L_2$ be two interpreted diagrammatic languages.  
Let $\Theta_1$ be a non-empty set of all well-formed expressions of $L_1$, such that there is at least one diagram $D$ in $\Theta_1$ and one entity $x$, such that $D$ is about $x$ in $L_1$. 
Let $\Theta_2$ be a non-empty set of all well-formed expressions of $L_2$, such that there is at least one diagram $D$ in $\Theta_2$ and one entity $x$, such that $D$ is about $x$ in $L_2$.

The function $m$ is a \emph{coarsening} from $\Theta_1$ (in $L_1$) to $\Theta_2$ (in $L_2$) iff
\begin{compactitem}
	\item $m$ is a function from $\Theta_1$ onto $\Theta_2$; and
	\item for all diagrams $D$ in $\Theta_1$ and all entities $x$: if $D$ is about $x$ in $L_1$, then  $m$($D$) is about $x$ in $L_2$; and
	\item for all diagrams $D_2$ in $\Theta_2$ and all entities $x$: if $D_2$ is about $x$ in $L_2$, then there is a diagram $D$ such that $D$ is about $x$ and $m$($D$) = $D_2$.
\end{compactitem}
\end{definition}

Coarsening functions map between two different diagrammatic languages, such that if a diagram in one language represents an entity, then it is possible to construct a diagram in the other language that also represents the entity.  \fn{Typically}, coarsening functions are \textit{directed} from a greater to a lesser level of detail; that is,  it is possible to map diagrams in a more detailed language to a diagram in a coarser language, but not the reverse.  Coarsening functions allow us to define a relationship \textbf{coarser than} between \textit{SD}s.

\begin{definition}
Let $D_1$ and $D_2$ be diagrams conforming to languages $L_1$ and $L_2$, respectively.
$D_2$ is \emph{coarser than} $D_1$ iff 
\begin{compactitem}
	\item there exists a function $m$ and sets of diagrams $\Theta_1$, $\Theta_2$ of $L_1$ and $L_2$, respectively, such that $m$ is a coarsening from $\Theta_1$ (in $L_1$) to $\Theta_2$ (in $L_2$) and  $m$($D_1$) = $D_2$; and
	\item there is no function $m'$ such that $m'$ is a coarsening from $D_2$ (in $L_2$)  to $D_1$ (in $L_1$).
\end{compactitem}
\end{definition}



\begin{figure}[htp]
\centering
	\includegraphics[scale=0.45]{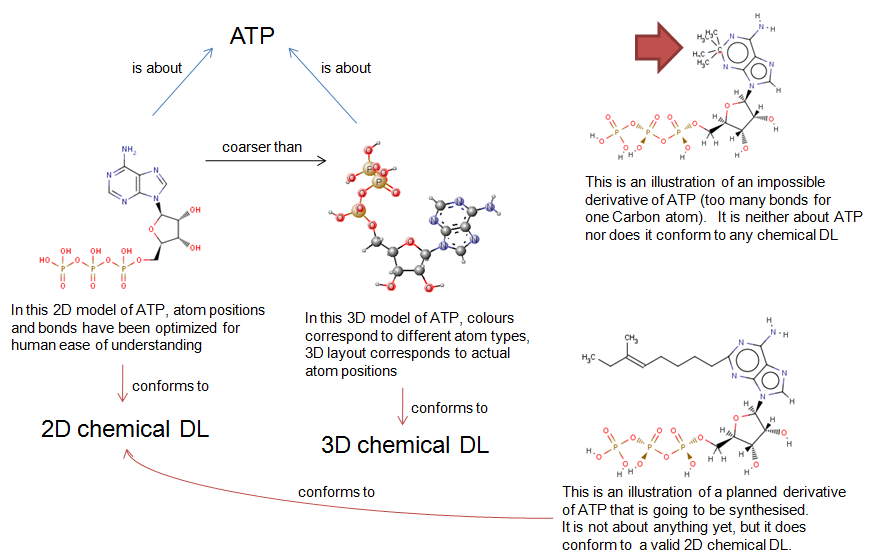}
\caption{Some examples of chemical diagrams and their relationships}
\label{fig:FigChemDiagRelationships}
\end{figure}

This is illustrated in Figure~\ref{fig:FigChemDiagRelationships}.

\section{Conclusion}

We have argued that the \textbf{is about} relationship is not enough to define \textit{CD}s, for two reasons.  Firstly, given the possibility of having several different \textit{CD}s corresponding to the same molecule, we see that distinguishing between different types of diagrams, which obey different representational syntaxes, is not possible using only distinctions in what the diagram \textbf{is about}.  Secondly, a challenge is posed in that \textit{CD}s may be used  validly to illustrate classes of molecules \textit{for which no instances exist}. The existential dependency expressed in IAO means that the IAO cannot, in its present form, allow for the inclusion of such non-referring information entities. 

We evaluated an approach based on parallel maintenance of IAO hierarchies with differing \textbf{is about} commitment.  While such parallel maintenance may be a scientifically-valid strategy in some scenarios, it is unable to express the fact that the same representational formalism (i.e., diagrammatic syntax) is used across the hierarchies. Of course, the diagrammatic syntax, if it is to be scientifically valid, must \textit{typically} represent entities which do exist. But the syntax allows for compositionality and it would be absurd to require the existence of instances for all the complex expressions obtained by composing the elements of the representational vocabulary.

We therefore propose the definition of structural diagrams such as chemical diagrams based on their syntaxes. Any diagram expressed in an interpreted diagrammatic syntax is a valid information content entity regardless of the existence of instances that the diagram \textbf{is about}; although the existence of such an instance may be an interesting property depending on the application scenario.

\bibliographystyle{splncs03}
\bibliography{C:/Work/Bibliography/mybib}
\end{document}